# ML-SPEAK: A Theory-Guided Machine Learning Method for Studying and Predicting Conversational Turn-taking Patterns

Lisa R. O'Bryan[1,2], Madeline Navarro[1], Juan Segundo Hevia[3], and Santiago Segarra[1]

[1]Department of Electrical and Computer Engineering, Rice University,

Houston, TX, USA

[2]Department of Psychological Sciences, Rice University, Houston, TX, USA

[3]Department of Computer Science, Rice University, Houston, TX, USA

**Author Note:**

This paper is awaiting peer review and has not yet been published

Funding for this project was provided by the Army Research Institute for the Behavioral and Social Sciences (Grant No. W911NF-22-1-0226).

The authors have no conflicts of interest to disclose

Code and data are available at the following link:

https://osf.io/brxk3/?view_only=bf194fb355a14c2d8a990718cb073f24

Correspondence concerning this manuscript should be addressed to Dr. Lisa O'Bryan, Rice University, Department of Electrical and Computer Engineering, MS-366, 6100 Main Street, Houston, TX 77005. Email: obryan@rice.edu, Phone: 713-385-9257

# Abstract

Predicting team dynamics from personality traits remains a fundamental challenge for the psychological sciences and team-based organizations. Understanding how team composition generates team processes can significantly advance team-based research along with providing practical guidelines for team staffing and training. Although the Input-Process-Output (IPO) model has been useful for studying these connections, the complex nature of team member interactions demands a more dynamic approach. We develop a computational model of conversational turn-taking within self-organized teams that can provide insight into the relationships between team member personality traits and team communication dynamics. We focus on turn-taking patterns between team members, independent of content, which can significantly influence team emergent states and outcomes while being objectively measurable and quantifiable. As our model is trained on conversational data from teams of given trait compositions, it can learn the relationships between individual traits and speaking behaviors and predict group-wide patterns of communication based on team trait composition alone. We first evaluate the performance of our model using simulated data and then apply it to real-world data collected from self-organized student teams. In comparison to baselines, our model is more accurate at predicting speaking turn sequences and can reveal new relationships between team members' traits and their communication patterns. Our approach offers a more data-driven and dynamic understanding of team processes. By bridging the gap between individual personality traits and team communication patterns, our model has the potential to inform theories of team processes and provide powerful insights into optimizing team staffing and training.

**ML-SPEAK: A Theory-Guided Machine Learning Method for Studying and**

**Predicting Conversational Turn-taking Patterns**

Organizations have long searched for insights into the relationship between team trait composition (e.g., personality, skill, ability) and team effectiveness (Bell et al., 2018; Kozlowski & Ilgen, 2018). Oftentimes, surveys are used to estimate whether an individual's traits are complementary to the trait composition of a given team, with the aim of producing a team that is both high-functioning and harmonious (Kichuk & Wiesner, 1997). Nevertheless, researchers still have a limited ability to predict the social, and work-related, interaction dynamics that will emerge from a team of a given composition. This is unfortunate since team interaction processes are a key intermediary for many team emergent states and outcomes (Hackman & Morris, 1975). Thus, improving the ability to predict team processes from team composition variables can provide greater control over the process of team development and training, increasing the likelihood that teams will display desirable outcomes.

## Importance of Communication in Teamwork

Team communication is defined as the exchange of information between team members (Marlow et al., 2018). Although communication can occur through both verbal and non-verbal channels, our focus here is on verbal communication. Communication is considered a key teamwork competency (Salas et al., 2018) and is an important facilitator of team outcomes, such as team coordination, decision-making, performance, and satisfaction (Marks et al., 2001; Marlow et al., 2018). For example, communication is associated with the distribution of information among team members (Salas et al., 2005), the improvement of team processes (Marks et al., 2001; Salas & Cannon-Bowers, 1997), and the development of team emergent states (Marlow et al., 2018). Previous research shows that, although communication is often positively associated with team outcomes, whether, or how much, communication benefits teamwork can vary according to the qualities of the communication being used (Marlow et al., 2018). Thus, studying the properties of the communication used by team members is important for understanding its connection to team outcomes.

**Patterns of Conversational Turn-taking**

Previous research has often focused on the qualities of communication used in teamwork. For example, Marlow et al. (2018) found that communication that functions by elaborating upon information is more important for enhancing team performance than other forms of communication, such as simply sharing information or knowledge (Marlow et al., 2018). However, in addition to the content of communication, there is emerging evidence that patterns of communication between team members play an important but understudied role in teamwork (Antone et al., 2020; McLaren & Spink, 2019; Sherf et al., 2018; Woolley et al., 2010). For example, communication network density (the proportion of dyadic interactions that occur out of all possible dyads) has been found to correlate with team member perceptions of team cohesion (i.e., remaining united in achieving a goal) (McLaren & Spink, 2019). In addition, the centralization of voice (the communication of work-related ideas) around dominant team members has been found to correlate with perceptions of less effective team member skill utilization and lower team performance, although voice centralization around more reflective team members was not associated with negative outcomes (Sherf et al., 2018). In general, higher variance in the number of speaking turns spoken across team members has been associated with poorer team outcomes (Li et al., 2019a; Woolley et al., 2010). For example, less inclusive patterns of conversational turn-taking have been found to correlate with a lower ability to identify expertise within the team (Haan et al., 2021) and, consequently, lower team performance. Even when communication behaviors occur in a one-to-all format (i.e., communication during in-person meetings that can be heard by all team members), dyadic patterns of interaction between team members can impact both individual and team-level outcomes (Humphrey & Aime, 2014; Humphrey et al., 2017). For example, (Schober & Clark, 1989) found that individuals within a group who are specifically addressed by another member gain a better understanding of the topic of discussion than other individuals who are present in the conversation but not directly interacting. Presumably, this is because the message is geared towards the addressee and they have the opportunity to develop a mutual understanding with the addresser. Such patterns of reciprocal turn-taking can also lead members of a dyad to develop more similar views of the topic being discussed (Fay et al., 2000; Schober & Clark, 1989).

**Traditional Studies of Team Communication**

Team processes, such as communication, can be strongly shaped by team composition (Bell et al., 2018). This is because a team member's individual characteristics can impact their speaking behaviors, and the combination of individual differences across team members can impact the speaking patterns that emerge. The link between individual characteristics and communication processes is often studied within the framework of the Input-Process-Output (IPO) model (Hackman & Morris, 1975). The term "inputs" encompasses team member characteristics (i.e., personality, demographics) as well as team attributes (e.g., team size) and environmental variables (e.g., type of task). Processes refer to team member behaviors and interactions, such as those related to communication, as well as coordination and cooperation (Kozlowski & Bell, 2003). Outputs include both objective and perceived outcomes of teamwork such as team performance and team satisfaction (Hackman & Morris, 1975). The IPO framework conceptualizes how team processes mediate the impact that team inputs have on team outputs.

Studies using these methods typically examine correlations between inputs, processes, and outputs. Data on communication or other team processes are often summarized at the end of a study or sampled at few points throughout a study through the use of self-, or peer-report methods or through the observation of these processes. These measures typically capture information on the frequency, quality, and/or type of information exchanged throughout teamwork (Bunderson & Sutcliffe, 2003; Kessel et al., 2012; Marlow et al., 2018). Using these methods, studies have found, for example, that the proportion of extroverts in a team (input) negatively correlates with task focus (process), team members' levels of social sensitivity (input) are positively correlated with the team's collective intelligence (output), and communication (process) positively relates to team performance (output) (Barry & Stewart, 1997; Marlow et al., 2018; Woolley et al., 2010). Such studies have provided a greater understanding of how different components of teamwork impact one another.

**A Need for a Mechanistic Understanding**

Although the IPO model is the most common framework for conceptualizing the connections between team inputs, processes, and outputs, its focus on correlations between summaries or snapshots of these measures render it incapable of providing a mechanistic understanding (Nicholson, 2012) of how the interactions between team members lead to the emergence of team processes and, ultimately, team outcomes (O'Bryan et al., 2020). For example, due to its focus on broad correlations between teamwork components, the IPO model has a limited ability to provide detailed explanations or predictions about how certain combinations of team inputs (e.g., team member personality traits, team size) result in specific team processes or emergent states (e.g., speaking time, leadership emergence, team cohesion). This limitation is particularly relevant to the examination of unique combinations of team inputs that have not been previously investigated. Related to this topic, psychological studies typically employ classical statistical models like linear regression, which may be insufficient for describing realistic, complex behavior (Benson & Campbell, 2007; Cucina & Vasilopoulos, 2005).

Teams can be considered complex systems (McGrath et al., 2000) in which team dynamics emerge from the interplay between team member characteristics, actions, and external factors, such as shifting task requirements (Kozlowski & Ilgen, 2018). Thus, team processes and outcomes are not necessarily equal to the sum of their individual components. During team communication, conversational turn-taking emerges from the interaction of many team members as they both influence, and are influenced by, the speaking behaviors of other members. This self-organized process is also shaped by the implicit "rules" of conversational turn-taking (Bonito, 2002; Kozlowski & Ilgen, 2018; Sacks et al., 1978; Stasser & Taylor, 1991) and can lead to the emergence of distinct speaking patterns. One such pattern, described by Parker (1988), is the ABA sequence. In this pattern, speaker A speaks, then speaker B, then speaker A again, with this pattern repeating for three or more turns. Parker (1988) and Stasser and Taylor (1991) refer to these ABA sequences as "floor" states and found that they operate under different implicit "rules" compared to other conversational states. These interaction patterns highlight a key limitation of traditional research approaches since summaries or snapshots of team processes fail to capture these fundamental conversation dynamics, rendering it impossible to understand how they emerge from the individual behaviors and interactions of team members. Without this understanding, it may be difficult to compose teams that will produce desired team processes (e.g., equal participation, centralization around one or a subset of team members, strong dyadic interactions) or develop interventions that can modify or improve team processes after they develop.

**Previous Models of Conversational Turn-taking**

Computational models can serve as powerful tools for studying team dynamics (McGrath et al., 2000). In particular, these models enable the development and testing of teamwork theories that fully embrace the complexities and interdependencies that arise from interactions between multiple team members (Harrison et al., 2007; Paoletti et al., 2021). When it comes to the study of communication patterns, Markov models are a common approach. Parker (1988) found that a second-order Markov model, which uses the previous two speakers to predict the next, significantly outperformed a first-order Markov model, which only considers the previous speaker. Predictions were not improved with higher-order models, which take into account greater numbers of preceding speaking turns. Instead, incorporating context-based conditional speaking probabilities resulted in the highest model performance. For example, researchers found that if a floor state (e.g., ABA) is currently in progress, this state is most likely to continue. If a floor state is interrupted by a third speaker, the most likely outcomes are that the previous floor is reestablished or a new floor is established with the new speaker. These patterns are more likely than group members transitioning to a non-floor state (i.e., unordered sequences of speaking turns between multiple team members). Building on the conversation model by integrating these context-based probabilities led to more realistic conversation patterns and more accurate predictions.

Developing upon the work of Parker (1988), Stasser and Taylor (1991) developed a conversation model (the SPEAK model) to investigate how both stable and state-based differences among speakers influence shifts between group-wide conversational states. Their model accounts for each member's baseline likelihood of initiating a speaking turn, as well as how the time since they last spoke affects their future speaking likelihood. After training their model on data from real groups of different sizes, they found that it successfully replicated the frequency of conversation state transitions displayed by these groups (e.g., the likelihood of transitioning between a floor and non-floor state). In addition to the model by Stasser and Taylor (1991), others have developed individual-based conversation models that choose each group member's actions based on a set of characteristics. For example, a model developed by Padilha and Carletta (2002) linked the probability distributions of a variety of verbal and non-verbal behaviors to broad traits possessed by the individuals, such as interactivity and talkativeness. In doing so, the researchers were able to examine the communication dynamics that emerge within groups with different trait compositions. However, a limitation of this method is that the relationship between individual characteristics and communication behaviors is predetermined by the model rather than learned from data. Accordingly, this type of model cannot provide new insights into the relationships between individuals' characteristics and their communication behaviors.

Finally, researchers have also used learning-based approaches to model conversation dynamics. These methods involve training algorithms on conversation data to predict future speaking patterns. For example, Basu et al. (2001) developed a model that, when trained on a specific dataset, learns both constant parameters reflecting each speaker's influence on others' speaking behaviors and pairwise transition probabilities between speakers. Their model successfully captured key communication patterns in both the simulated and real datasets they trained it on. However, a shortcoming of this approach is that the learned parameters are specific to the team members in the training datasets. Thus, the model can learn past patterns of influence within a team but cannot learn the mechanisms underlying this influence (i.e., the role of team member trait profiles) and apply this understanding to new teams in order to predict their interactions.

**Personality-driven Machine Learning**

Machine learning models learn to make predictions by empirically capturing relationships between recorded traits and associated outcomes. The success of learning-based computational models for extracting complex patterns from data has led to their application towards personality-focused analysis (Bleidorn & Hopwood, 2019). Such modern learning-based tools align with interest in the predictive ability of personality traits (Cuperman & Ickes, 2009). However, most works on machine learning for personality assessment apply models that are either interpretable yet lack expressivity (i.e., the level of complexity of the functions or relations that can be learned is low) or powerful but considered "black box" approaches (Stachl et al., 2020). For example, while non-linear machine learning models such as neural networks show major empirical success, their parameters tend to be uninterpretable, obtained to optimize predictive ability without revealing how personality traits relate to outcomes (Yarkoni & Westfall, 2017).

Indeed, neural networks typically possess too many indicators for systematic examination. However, models can be simplified by additional assumptions, which can be domain-dependent (Karniadakis et al., 2021). For example, physics-informed neural networks require that models follow universal physical laws, not only simplifying the computational learning task but also providing a more conceptually relevant model (Ghaseminejad & Uddameri, 2020; Guest et al., 2018; Praditia et al., 2020). The incorporation of domain knowledge is thus well-founded in data science and also stipulated for personality-focused machine learning (Bleidorn & Hopwood, 2019; Mahmoodi et al., 2017; Shmueli, 2010; Stachl et al., 2020). In the context of conversational turn-taking, prior assumptions can come in the form of stable rules of interactions from which common patterns of interaction are known to emerge (Parker, 1988; Stasser & Taylor, 1991). Still, these rules allow for individual variation across team members that can be important for influencing group outcomes (Couzin et al., 2002; Jolles et al., 2017). By building models that take into account basic interaction rules, as well as individual variation around these rules, researchers can predict social outcomes such as cohesion, coordination, and decision-making (Couzin et al., 2002; Passino & Seeley, 2006).

## Computational Methods

We present our computational model of conversational turn-taking which serves as a statistical model of team member speaking behaviors dependent upon their individual traits. We extend and apply a learning-based approach to an existing conversational model (the SPEAK model) that describes verbal communication patterns within small groups (Stasser & Taylor, 1991; [citation to be added after review]). By exploiting potent deep learning tools for parameter fitting, we can relate individual differences to speaking behaviors by estimating how model parameters vary across team members based on their individual traits. Thus, we name our model the *ML-SPEAK* model. We first introduce the conversational model, then we describe our computational approach to learning behavioral parameters from team conversations.

We consider a simple parametric conversational model consisting of two key notions: i) the relative likelihood $\pi_i$ that team member $i$ speaks on a given turn independent of their speaking history, and ii) the strength $d_i$ of the effect that an individual's current speaking turn has on their likelihood of speaking on subsequent turns [citation to be added after review]. More precisely, the likelihood $\ell_i(t)$ that member $i$ speaks at turn $t$ is given by

$$\ell_i(t) = \left\{ \begin{array}{cc} 0, & \text{if } t_i^{\text{last}} = t - 1 \\ \pi_i + d_t e^{-0.5\left(t - t_i^{\text{last}}\right)}, & \text{otherwise} \end{array} \right\}, \tag{1}$$

where $t_i^{\text{last}}$ denotes the last turn on which member $i$ spoke. The negative exponential form in $\ell_i(t)$ reflects the asymptotically decaying effect due to the most recent speaking turn as more turns occur, that is, as $\left(t - t_i^{\text{last}}\right)$ increases. This encodes the natural assumption that an individual speaking several turns in the past is inconsequential to their likelihood of speaking next. We denote the second summand $d_i e^{-0.5\left(t - t_i^{\text{last}}\right)}$ the memory term or memory function since this term encodes how previous speaking behavior affects the likelihood of speaking next. Furthermore, a speaker's turn ends only when another member speaks, so no member can speak on two consecutive turns, hence the likelihood $\ell_i(t)$ being set to zero if member $i$ has just spoken. Lastly, denoting by $N$ the total number of team members, the likelihoods $\ell_i(t)$ are normalized to sum up to 1 to yield valid probabilities $p_i(t)$ as follows

$$p_i(t) = \frac{\ell_i(t)}{\sum_{j=1}^{N} \ell_j(t)},$$

with the speaker at turn $t$ being drawn from this probability distribution across team members. Thus, the conversational behavior of each individual $i$ within this conversation model is entirely described by the dyad of parameters $\left(\pi_i, d_i\right)$.

Given these parameters for every team member, the model provides a well-defined stochastic process to generate conversations by the team. More precisely, let the dyads $\left(\pi_1, d_1\right), \ldots, \left(\pi_N, d_N\right)$ correspond to $N$ team members. To sample the next speaker at turn $t$, we first compute the value $\ell_i(t)$ for each member $i$ given the last speaking turn $t_i^{\text{last}}$, resulting in a set of likelihoods $\ell_1(t), \ldots, \ell_N(t)$. Then, we compute the probability distribution of the next speaker as $p_1(t), ..., p_N(t)$, from which we draw the team member who speaks next. This process is repeated for as many turns as occur in the conversation, yielding a sequence of speaking turns for the team of interest.

Unlike the model by Stasser and Taylor (1991), which sets speaker behavior based on team-level parameters (i.e., one parameter that sets a proportional drop in baseline speaking probability from one speaker rank to the next and one value of $d$ for all team members), our model learns individual differences in speaking behavior. We use a set of teams for training for which we have both the traits of the team members and their conversation history. To associate individuals' traits with speaking behaviors, we feed each team member's measured traits through a neural network to predict each speaker's dyad of parameter values $\left( \pi_i, d_i \right)$. The parameters of the neural network are fit to maximize the likelihood that the observed conversations occur given the team member speaking probabilities. In other words, we train our neural network to learn a map from an individual's traits onto individual model parameters $\left( \pi_i, d_i \right)$ such that the probability of generating the observed conversations is maximized. Specifically, we consider a two-layer feedforward neural network which takes as input a set of individual traits (e.g., $\left( a_i, b_i \right)$), and we set the number of neurons in the first layer to equals the number of traits. The second layer contains ten neurons that output a transformation of the traits, followed by the output layer of only two neurons returning dyad $\left( \pi_i, d_i \right)$. Once the neural network has been trained, any new trait values can be provided, and the neural network will output the predicted $\left( \pi_i, d_i \right)$ for that individual. Consequently, given the trait values of all the members in a team of interest, we can use the trained neural network to estimate their parameters $\left( \pi_i, d_i \right)$ and then follow the stochastic procedure previously described to generate realistic conversations for the team. Importantly, we can do this even in the *absence of any observed prior conversation between the team members*.

## Overview of Studies

We propose a learning-based conversation model, described in the Computational Methods section, that relates individual differences to team communication patterns. In Study 1 we test our model's performance on synthetic data in terms of its ability to predict specific speaking sequences. In addition, we compare the performance of our conversation model with other baseline models, enabling us to establish how our model compares to others in terms of predictive ability and gain insight into the value of considering individual differences when making such predictions. In Study 2, we again test our model on synthetic data while systematically varying model components and synthetic data properties, such as conversation length, group size, and properties of turn-taking sequences. This work enables us to determine how model performance is impacted by these variables which can inform decisions regarding the suitability of different datasets for use with our model. In Study 3, we apply our model to real-world data from student teams interacting virtually during semester-long team-based projects to test the association between team member personality traits and their speaking behaviors. This enables us to explore which traits or combinations of traits (if any) are most predictive of individual communication patterns, enabling us to demonstrate the utility of our approach using real data. We also compare the performance of our highest-performing model to the baselines models we tested in Study 1. This enables us to not only calculate the performance advantage of our model compared to other baselines but also assess the additional advantage associated with learning individual differences in speaking behaviors. In the Methods section of each study, we report all data exclusions (if any), all manipulations, and all measures used in the study. As our study focuses on the development of novel methodological approaches and exploratory analysis, this study was not preregistered.

<div align="center">**Transparency and Openness**</div>

The real-world data used in our study and our model code have been made publicly available through the Open Science Framework and can be accessed at

https://osf.io/brxk3/?view_only=bf194fb355a14c2d8a990718cb073f24.

<div align="center">**Study 1**</div>

In Study 1, we use synthetic data to test the predictive performance of our conversation model relative to other baseline models, enabling us to establish how our model compares to other models and the value that our model's consideration of individual differences plays in this predictive ability. We also demonstrate the ability of our model to learn the relationships between individual traits and model parameters by visualizing both the input functions used to generate these relationships in the synthetic data and the associated output functions learned by our model.

**Design and Method**

*Procedure*

All data used in Study 1 were synthetic. For each experiment, we consider 20 teams for training, five teams for validation, and five teams for testing. All teams consisted of five team members and produced one 600-turn conversation. In order to generate the synthetic conversations for the teams, we assigned each team member two individual characteristics (modeling generic personality traits), which we denote $a_i$ and $b_i$, generated uniformly at random between 0.1 and 1. To generate a speaker's parameter dyad $\left( \pi_i, d_i \right)$ from their traits $\left( a_i, b_i \right)$, we set $\pi_i = \sqrt{a_i}$ and $d_i = \frac{15}{2} \left( \frac{e^{-2b_i}}{e^{-0.2} - e^{-2}} - e^{-2} + \frac{1}{3} \right)$. We choose these functions to be monotonic, and the scales are chosen so that generated conversations tend to have similar speaking patterns to real team conversations. These functions encode universal relationships between the personality trait values of an individual $\left( a_i, b_i \right)$ and the parameters $\left( \pi_i, d_i \right)$ determining their turn-taking behavior.

Using the above settings, we generated 20 "data trials"- independent sets of training data made up of randomly selected trait values for each team member and new conversations generated from each team members' associated $\pi$ and $d$ values. A single validation dataset and a single testing dataset were used for all 20 data trials. We trained our *ML-SPEAK* model separately on each of the 20 training datasets, validating and testing its performance on the same validation and testing set each time, resulting in 20 performance values (See Analysis section for details of testing). In addition to our standard *ML-SPEAK* model, we trained and tested five model variants on the same training and testing datasets, allowing us to contextualize our model's ability to predict the speaking turn sequences in the test datasets. The five model variants used for comparison are described below.

1.  *Same traits*: For this model, we set all individual trait values to 0.5 ($a_i = b_i = 0.5$ for all individuals *i*) in the training and validation datasets. We then trained our model on these data and tested model performance using the original test dataset. This results in our model learning the same dyad values $\left( \pi_1, d_1 \right) = \left( \pi_2, d_2 \right) = \cdots = \left( \pi_5, d_5 \right)$ for all five team members of each team within a given data trial. However, for each data trial, our model still tunes the ratio of learned $\pi$ and $d$ values according to the best fit for each training dataset. By comparing our standard model to the *same traits* model, we can determine the added value of considering inter-individual variation in speaking behaviors in predicting speaking turn sequences.

2. *Same traits, no memory*: For this model, we set all trait values to 0.5 ($a_i = b_i = 0.5$ for all

   individuals $i$) in the training and validation datasets. We then used these data to train a

   version of our *ML-SPEAK* that does not include the memory function (i.e., $d_i = 0$ for all

   team members) and only takes into account team members' baseline likelihoods of

   speaking $\pi_i$. Accordingly, all team members have the same predicted speaking likelihood

   (with the caveat that the speaker who spoke last will not speak on the current turn). We

   tested model performance using the original test data. Comparing our standard *ML-SPEAK*

   model to the *same traits, no memory* model enables us to determine the added value of

   considering individual variation in speaking behaviors as well as the role of the memory

   function.

3. *Randomized traits*: For this model, we randomized team member trait values $a_i$ and $b_i$

   (separately for each trait) within each team in the training and validation datasets. We then

   trained our standard *ML-SPEAK* model on these randomized data and tested model

   performance using the original test dataset. By comparing our standard model to the

   *randomized traits* model, we can evaluate the added value of taking into account

   meaningful inter-individual variation in speaking behaviors (i.e., training the model on

   traits that actually played a role in generating variation in communication behaviors).

4. *Linear Regression*: For this model, we trained a linear regression model on the

   relationships between the two trait values assigned to each participant $\left( a_i, b_i \right)$ and the total

   number of speaking turns they displayed in the training dataset. We then used this model to

   predict the number of speaking turns expected of team members in the test datasets, based

   on their individual traits. These values were then normalized to sum to 1 within each team,

   representing the predicted speaking proportions of these team members. We used these values as $\pi_i$ values in our standard *ML-SPEAK* model without the memory component ($d_i = 0$) and then tested model performance using the original test dataset. By comparing our model to the *linear regression* model, we can evaluate the performance of our model against a common method of assessing the relationship between team member traits and team communication behaviors (i.e., regressing team member speaking proportions on personality traits) to determine whether our model provides a predictive advantage.

5. *SPEAK*: For this model, we trained a linear regression model on the relationships between the two trait values assigned to each participant $\left( a_i, b_i \right)$ and the total number of speaking turns they displayed in the training dataset. We then used this model to predict the number of speaking turns expected of team members in the test datasets, based on their individual traits. We used these values to rank speakers in the test groups from 1 to 5, with 1 indicating the most talkative team member and 5 being the least talkative team member. Based on these ranks, we then assigned these team members standardized $\pi_i$ values according to the original model by Stasser and Taylor (1991). More specifically, we let $\pi_i = \dfrac{r^i}{\sum_{j=1}^{5} r^j}$ with $r = 0.7$ denote the value for the speaker with rank $i$. Following the original model developed by Stasser and Taylor (1991), we let the memory component have the same weight $d_i$ for all speakers. We choose $d_i = 2.26$ such that the ratio of the median $\pi_i$ and median $d_i$ values equaled that in our *ML-SPEAK* model. These values were used as $\left( \pi_i, d_i \right)$ values in our standard *ML-SPEAK* model, and we tested model performance using the original test data set. By comparing our standard *ML-SPEAK* model to the *SPEAK* model, we can determine the added value of our approach to learning the associations between individual traits and the values of $\left( \pi_i, d_i \right)$ directly from the data.

*Analysis*

We used each trained model described above to calculate the log-likelihood (i.e., the loss value) of observing the sequences of speaking turns observed in the testing dataset where smaller loss values indicate better predictive performance. This loss value is calculated by using each model to determine the probability of selecting the true speaker at each speaking turn. For a single conversation containing $T$ turns, the loss is computed as

$$Loss = -\sum_{t=1}^{T} \log\left( p_{i(t)}(t) \right), \tag{2}$$

where $i(t)$ denotes the true speaker at turn $t$ and $p_{i(t)}(t)$ denotes the probability that the model under study assigned to the event that $i(t)$ speaks at turn $t$. We calculated the loss for each model for each data trial, resulting in 20 data points for each model.

While the conversations in Studies 1 and 2 are generated from a synthetic set of parameters $\left( \pi_i, d_i \right)$ for every team member, real-world conversations have no such parameters. Thus, when working with real-world data, we cannot compare the $\pi$ and $d$ values learned by the model to ground truth data. In order to promote comparison of model performance across Studies 1 and 3, we instead measure model performance by comparing its loss to that of a baseline model that cannot learn from personality traits (the *same traits* model). Specifically, we subtracted the loss values obtained by the *same traits* model for a given data trial from the loss values obtained by our model for that same data trial. Thus, a loss difference less than zero indicates that our model performed better than a model that did not consider individual variation in trait values. By extending this approach to all baseline modes, we used a Kruskal-Wallis test to test for differences in performance across model types. If differences were found, we used a Wilcoxon multiple comparisons test to determine which of the models significantly differed from one another in their performance.

We visualized the functions we used to generate the relationship between individuals' traits and their $\pi$, $d$ values in our synthetic data as well as the relationships our model learned after being trained on these data. We also visualized overall speaking likelihood $\ell_i\left(t_i^{\text{last}} + 2\right) = \pi_i + d_i e^{-0.5}$, that is, the highest speaking likelihood that an individual $i$ can have, which occurs at the earliest instance that member $i$ can speak after the last turn. The visualizations of the learned output functions represent the average relationship learned by the *ML-SPEAK* model across all 20 data trials.

**Results**

Model type had a significant effect (Kruskal-Wallis chi-squared = 95.05, p < 0.001) on the ability to predict speaking turn sequences (i.e., loss value), with all models performing significantly differently from one another (Figure 1). Our ML-SPEAK model was the highest-performing model, followed by the *SPEAK*, *randomized traits*, *linear regression*, and *same traits, no memory* models, in that order. The models that did not include the memory component (the *linear regression* and *same traits, no memory* models) performed significantly worse than all models that did include this component. Within the model categories that either did or did not include memory, the models that fit model parameters based on meaningful differences between individuals (Memory models: *ML-SPEAK*, *SPEAK*, No Memory Models: *linear regression*) performed better than the models that did not. Finally, we see that our *ML-SPEAK* model, which learns $\pi$ and $d$ values for each team member from the data, performed better than the *SPEAK* model which forces a speaking hierarchy on team members and includes the same $d$ value for all individuals.

Our model performed well in learning the input functions used to generate the relationships between an individual's trait dyad, $\left( a_i, b_i \right)$, and their model parameters, $\left( \pi_i, d_i \right)$ as well as their overall speaking likelihood $\ell_i\left( t_i^{last} + 2 \right)$ (Figure 2). One notable difference is that the input function controlling the shape of $\pi$ had a convex shape while the learned output function had a concave shape. Nevertheless, both demonstrate a positive relationship between $a_i$ and $\pi_i$.

**Figure 1**

*Model Performance Comparisons for Synthetic Data*

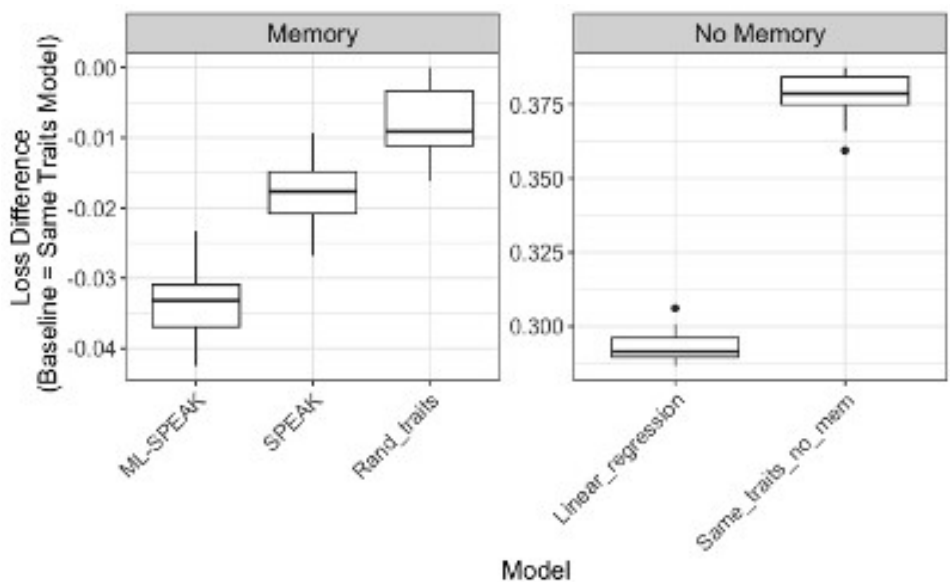

*Note.* "Loss Difference" results represent the difference in loss between each model and the *same traits* model with lower values indicating better performance. "ML-SPEAK" corresponds to our proposed model, while the baselines listed in the Procedure subsection for Study 1 are denoted in the figures by "Same_traits", "Same_traits_no_mem", "Rand_traits", "Linear_Regression", and "SPEAK", respectively.

**Figure 2**

*Trait Relationships Learned from the Synthetic Data*

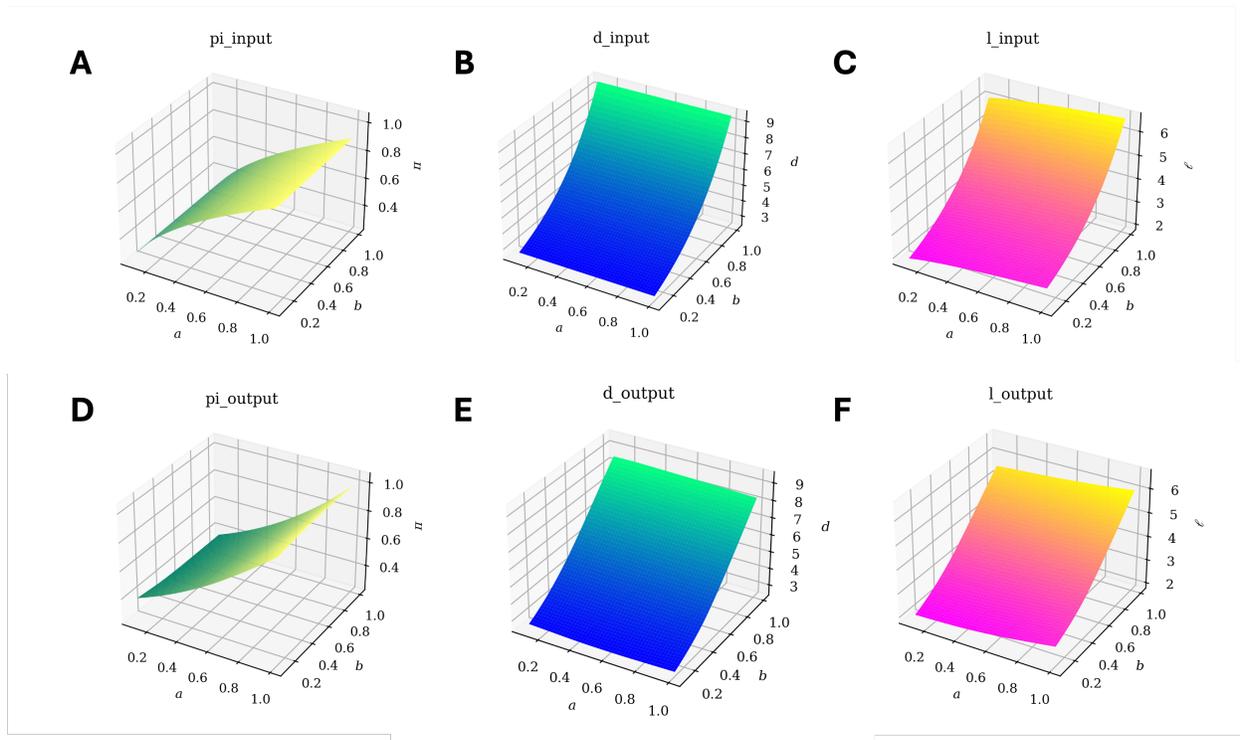

*Note.* A and B represent the functions used in Study 1 to control the relationships between an individual's trait dyad $\left( a_i, b_i \right)$, and their model parameters, $\left( \pi_i, d_i \right)$. C represents the relationship between an individual's traits and their overall speaking likelihood $\ell_i \left( t_i^{\text{last}} + 2 \right)$ on the turn falling two turns after they just spoke (i.e., the earliest they can initiate another speaking turn). D-F represent the associated functions learned by our *ML-SPEAK* model after training it on team member traits and speaking sequences.

**Discussion**

The results from Study 1 highlight the importance of learning individual differences in speaking behaviors and leveraging the memory function to predict patterns of conversational turn-taking when analyzing conversational data that includes both individual variation in turn-taking patterns and strong memory-dependent turn-taking behavior. Although it is not surprising that our model performed better than other baselines, which were less closely aligned with the processes used to generate the synthetic data used in this study, these results serve as a valuable point of comparison for results obtained when working with conversation data for which the processes generating the conversations are unknown (as in real-world data), which we explore in Study 3. Study 2 extends these findings by assessing how our model performs when there is a mismatch between the model used for inference and the processes underlying the turn-taking data (i.e., the presence of individual differences and/or strong turn-taking behavior).

Our results from Study 1 also demonstrate the ability of our model to learn existing relationships between individual traits and model parameters, providing confidence in results obtained when the input functions controlling these relationships are unknown, as in real-world data.

## Study 2

In Study 2, we systematically varied model components as well as properties of the synthetic data to assess how these variables impact model performance. This enables us to assess the value of different model components and provides insight into how properties of different datasets may impact model performance.

### Design and Method

#### *Procedure and Analysis*

All data used in Study 2 were synthetic. Like Study 1, all experiments used 20 teams for training, five teams for validation, and five teams for testing. For all experiments, except the Group Size experiment, teams consisted of five team members. Each team produced one 600-turn conversation in all experiments except the Conversation Length experiment.

We generated 20 data trials for each condition within each experiment. As in Study 1, training data was independently generated for each data trial while the same datasets were used to validate and test all data trials within a given experiment, with the exception of the Trait Complexity experiment. We used two individual characteristics for all experiments. The type of function relating individual traits $\left( a_i, b_i \right)$ with the values of $\pi_i$ and $d_i$ differed across experiments. We systematically varied different aspects of the synthetic datasets (conversation length, group size, type of function relating individual traits with $\pi$ and $d$) and model (the presence of the memory function, the ability to leverage individual differences) in order to assess their effects on model performance. In the following experiments, we used each trained model to calculate the log-likelihood (i.e., the loss value) of observing the sequences of speaking turns observed in the testing datasets and compared these loss values across the different conditions within each experiment.

**Data / Model Type.** Real conversational turn-taking behavior is assumed to be time-dependent (Stasser & Taylor, 1991; [citation to be added after review]), which is reflected in our model via the parameter $d$. However, the processes that generate patterns of conversations are unknown in real-world data. Thus, our second study explores how the match between the processes underlying conversational data and the parameters taken into account by the model affects model performance. Similar to Study 1, we independently generated training data for each data trial. Furthermore, for each trait dyad $( a, b )$, we obtained a parameter dyad $( \pi, d )$ by using the complex, uncorrelated function (defined in the Trait complexity subsection below). Using this approach, we then constructed three types of datasets. First, we generated conversations from the parameters $(\pi, d)$ as usual, which we call our *Memory* dataset. Second, we generated a *No Memory* dataset by letting $d = 0$ for all individuals and generating conversations using only the baseline speaking likelihood $\pi$ for each team member. This inhibits the presence of marked dyadic turn-taking behavior in the conversation data. Third, we used the original values for $d$ as in the *Memory* dataset, but the baseline speaking likelihood was set to be constant across all speakers, that is, $\pi = 0.1$ for all individuals, which we term *Same-Pi*. Thus, the conversation data reflects individual differences in turn-taking behaviors but not baseline likelihoods of speaking. Analogously, we compare three types of models. First, corresponding to the *Memory* dataset, we train our standard *ML-SPEAK* model, which learns individual differences in $\pi$ and $d$ from traits $a$ and $b$. Second, we train a *No Memory* model with no memory component (i.e., $d = 0$ for all individuals), which can only learn individual differences in $\pi$ and does not take into account a greater tendency to speak after having recently spoken. Third, we consider a *Same-Pi* model that emphasizes the memory component by learning an individual value $d_i$ for team member $i$, but assigns every team member the same value of $\pi = \pi_1 = \cdots = \pi_5$. We ran all 9 data / model combinations and used a Kruskal-Wallis test to test for any differences in performance across model types within each dataset type. If differences were found, we used a Wilcoxon multiple comparisons test to determine which of the model types significantly differed from one another for each dataset type.

**Trait complexity.** Presumably, simple functions (i.e., linear relations) controlling the relationships between individual traits and model parameters should be easier to learn than more complex relationships. Thus, we generate functions of varying complexity to determine how the complexity of the relationship to be learned affects the performance of our model in predicting speaking sequences. To generate a speaker's dyad $(\pi_i, d_i)$ from their traits $(a_i, b_i)$, we define two pairs of functions: one simple $f_{simple}$, $g_{simple}$ and one complex $f_{complex}$, $g_{complex}$. For the first, simpler pair, we let $f_{simple}(x) = x$ and $g_{simple}(x) = \frac{5}{3}(5x + 1)$. For the more complex pair, we set

$f_{complex}(x) = \sqrt{x}$ and $g_{complex}(x) = \frac{15}{2}\left(\frac{e^{-2x}}{e^{-0.2} - e^{-2}} - e^{-2} + \frac{1}{3}\right)$. From these functions, we generate

$(\pi_i, d_i)$ from $(a_i, b_i)$ with a relationship that is either simple via $f_{simple}$ and $g_{simple}$ or complex

via $f_{complex}$ and $g_{complex}$.

We further consider three cases: $\pi_i$ and $d_i$ are uncorrelated, $\pi_i$ and $d_i$ are positively correlated, and $\pi_i$ and $d_i$ are negatively correlated. In particular, we consider the following six conditions: (1) for simple and uncorrelated $\pi_i$ and $d_i$, we let $\pi_i = f_{simple}(a_i)$ and $d_i = g_{simple}(b_i)$;

(2) for simple and positively correlated $\pi_i$ and $d_i$, we let $\pi_i = f_{simple}(0.5(a_i + b_i))$ and

$d_i = g_{simple}(0.5(a_i + b_i))$; (3) for simple and negatively correlated $\pi_i$ and $d_i$, we let

$\pi_i = f_{simple}(0.5(a_i + b_i))$ and $d_i = g_{simple}(1.1 - 0.5(a_i + b_i))$; (4) for complex and uncorrelated

$\pi_i$ and $d_i$, we let $\pi_i = f_{complex}(a_i)$ and $d_i = g_{complex}(b_i)$; (5) for complex and positively correlated

$\pi_i$ and $d_i$, we let $\pi_i = f_{complex}(0.5(a_i + b_i))$ and $d_i = g_{complex}(0.5(a_i + b_i))$; and (6) for complex

and negatively correlated $\pi_i$ and $d_i$, we let $\pi_i = f_{complex}(0.5(a_i + b_i))$ and

$d_i = g_{complex}(1.1 - 0.5(a_i + b_i))$.

We independently generated new data for each data trial (i.e., different values of $a$ and $b$

for the different team members). For each condition we generated $\pi$ and $d$ based upon the specified relationship. In other words, we used the same values of $a$ and $b$ for all training groups across conditions, but $\pi$ and $d$ differ across conditions according to the trait function. We tested all six conditions listed in the previous paragraph. Training, validation, and testing data all had the same *a* and *b* values across all trait complexity types, but, again, $\pi$ and *d* were calculated using the trait function specific to a given condition. Therefore, the conversations in the training, validation, and testing datasets differ across conditions. We used a Wilcoxon Rank Sum test to test for model performance differences when trained on data generated from the simple versus complex functions. Within each function type we also used a Kruskal-Wallis test to test for differences in model performance when trained on the negatively correlated, positively correlated, and uncorrelated traits.

**Conversation length.** We expected the amount of data available for training to impact model performance, though the amount of data needed for sufficient performance remained unclear. To gain insight into this topic, we tested ten conditions with synthetic conversations consisting of 50-500 turns in increments of 50. For each data trial, we first generated a 500-turn dataset. Then, keeping everything else consistent, we cropped the conversation to each of the new lengths. We compared results between conditions using either complex functions generating $\pi$ and *d* values from individual traits that resulted in a negative correlation between $\pi$ and *d* ('complex, negative correlation') or simple functions that resulted in uncorrelated values of $\pi$ and *d* ('simple, uncorrelated'). We used a Wilcoxon Rank Sum test to test for performance differences between models trained on simple and complex traits. Within each model type, we also used a Kruskal-Wallis test to test for differences in performance between conversation length categories. If differences were found, we used a Wilcoxon multiple comparisons test to determine which of the conversation length categories significantly differed from one another.

**Group size.** We expected group size to have the potential to impact model performance since larger group sizes may involve more complex turn-taking interactions between group members and may also result in fewer speaking opportunities for team members. To gain insight into this topic, we tested four group sizes while keeping the total number of team members across teams equal to 120. Our conditions consisted of 12 teams of 10 team members each, 15 teams of 8 team members each, 20 teams of 6 team members each, and 30 teams of 4 team members each. All 120 individuals were held consistent across data trials. Across different data trials, team members were randomly assigned to different teams. Like the conversation length experiment, we compared results between conditions that used either a complex, negative correlation function to generate $\pi$ and $d$ or a simple, uncorrelated function. We used a Wilcoxon Rank Sum test to test for performance differences between models trained on simple and complex traits. Within each model type, we also used a Kruskal-Wallis test to test for differences in performance between group size categories. If differences were found, we used a Wilcoxon multiple comparisons test to determine which of the group size categories significantly differed from one another.

**Results**

     **Data / Model Type.** The type of model used for inference significantly impacted model performance across the different data types (Data type = Mem: Kruskal-Wallis chi-squared = 52.459, p < 0.001, Data type = NoMem: Kruskal-Wallis chi-squared = 42.959, p < 0.001, Data type = SamePi-Mem: Kruskal-Wallis chi-squared = 52.459, p < 0.001). Within each data type, all models performed significantly differently from one another (Figure 3A). We first focus on the data type that represented individual differences in both speaking likelihoods and the effect of memory ("Mem"), which is presumably the most representative of real conversations, as well as the data type that represented a single speaking likelihood for all individuals and individual differences in the effect of memory ("SamePi"). Within these two data types, the models that learned individual differences in both speaking likelihoods and memory parameters ("Mem")

performed the best, followed by the models that learned one speaking likelihood for all individuals but individual-specific memory parameters ("SamePi"), followed by the models that only learned individual speaking likelihoods and no effect of memory ("NoMem"). Within the data type that represented individual differences in speaking likelihoods and no effect of memory ("NoMem"), the model that learned individual differences in speaking likelihoods and no effect of memory ("NoMem") performed best, followed by the model that learned individual differences in both speaking likelihoods and memory parameters ("Mem"), and then the model that learned one speaking likelihood but individual-specific memory parameters ("SamePi").

**Figure 3**

*Results for Trait Complexity and Data / Model Type*

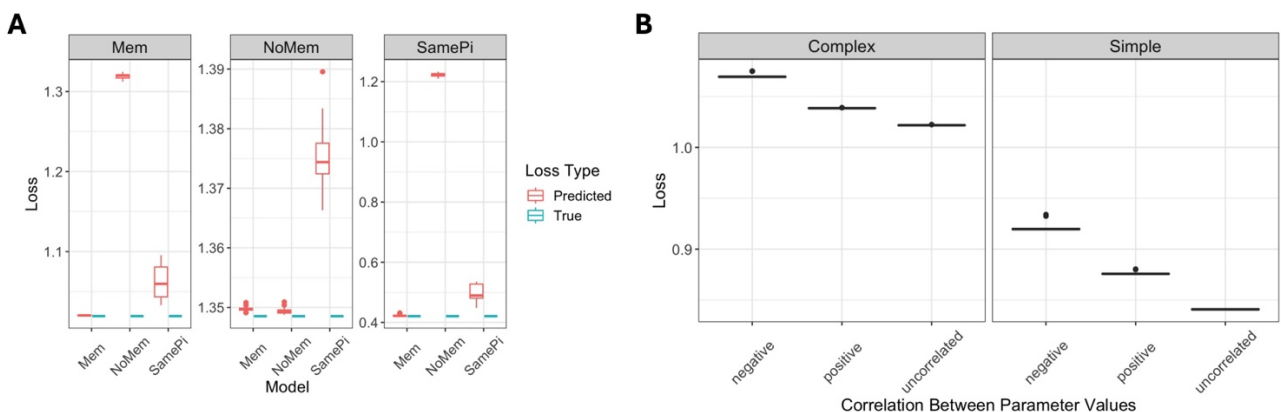

*Note.* The *x*-axis labels in A correspond to the type of model used, while the panels denote the type of dataset each model was trained on. "True" refers to the Loss obtained by using the true pi and d values. "Predicted" refers to the Loss obtained by using the pi and d values learned by the model. The panels in B represent the complexity of the function used to generate the relationships between individual trait values and model parameters ($\pi, d$). The *x*-axis labels in B correspond to the correlation between these parameter values.

**Figure 4**

*Results for Conversation Length*

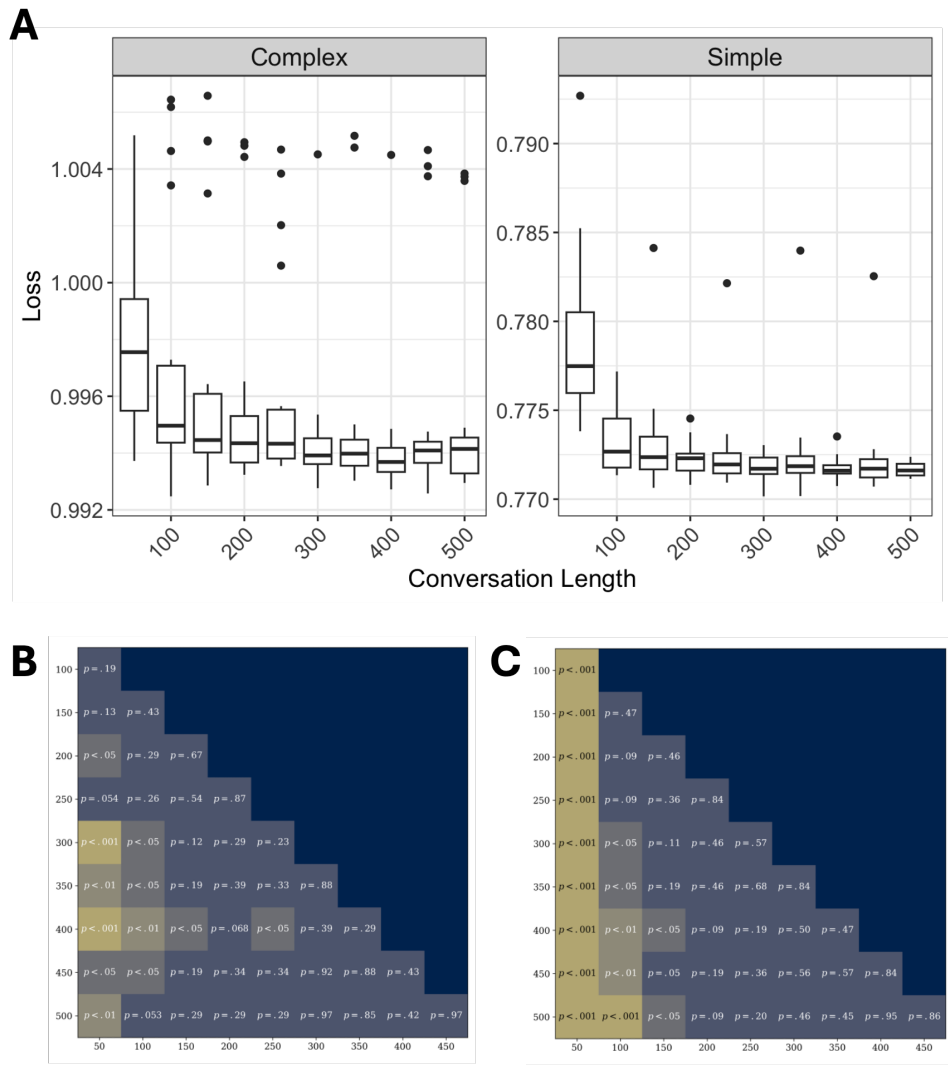

*Note.* "Simple" and "Complex" refer to the complexity of the function used to generate the relationships between individual trait values and $(\pi, d)$. The *x*-axis of A represents the length of the conversations for each team in the training dataset. In B and C, rows and columns represent the number of turns in each conversation by each team in the training datasets. The p-values and associated color shades correspond to the results of the Wilcoxon multiple comparisons test with lighter values indicating greater differences between model results.

**Figure 5**

*Results for Group Size*

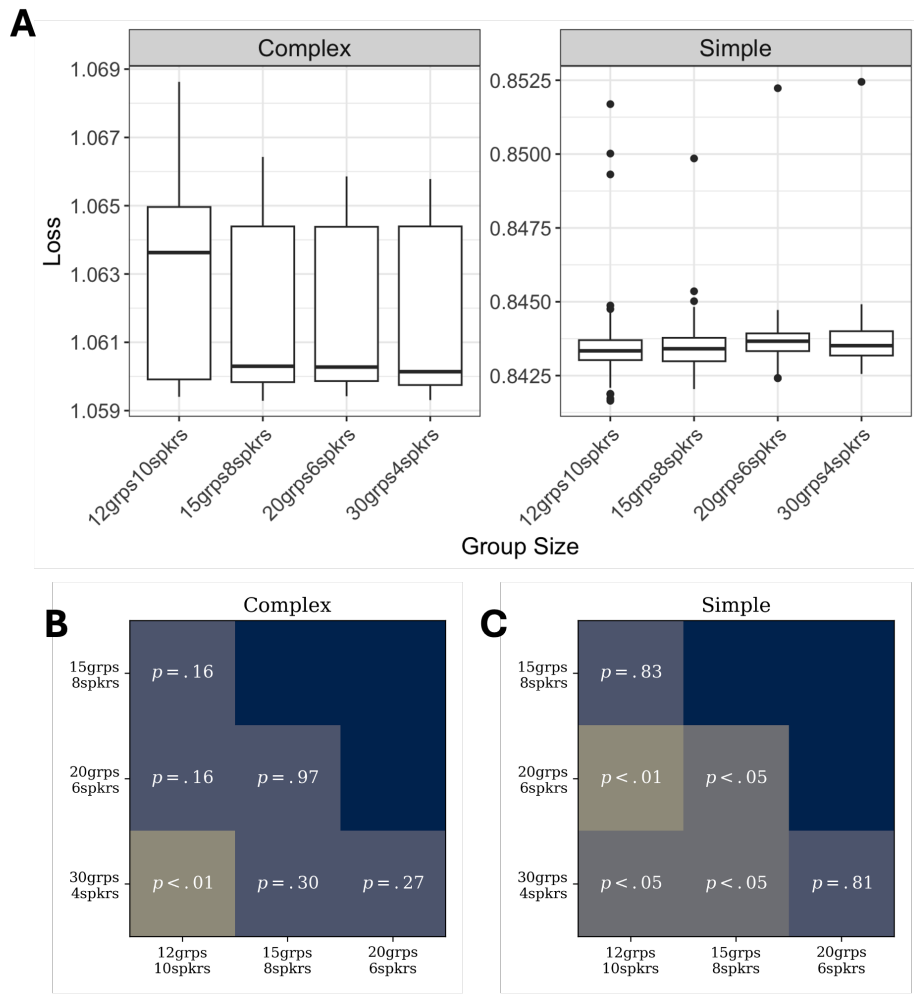

*Note.* "Simple" and "Complex" refer to the complexity of the function used to generate the relationships between individual trait values and $(\pi, d)$. The *x*-axis of A represents the number of groups and number of team members per group in the training dataset. In B and C, rows and columns represent the number of groups and number of team members per group in the training datasets. The p-values and associated color shades correspond to the results of the Wilcoxon multiple comparisons test with lighter values indicating greater differences between model results.

**Trait complexity.** Comparisons of model performance across different functions mapping $(a, b)$ to $(\pi, d)$ are shown in Figure 3B. When considering raw performance, models trained on simple traits performed significantly better than models performed on complex traits (W = 3600, p < 0.001). For models trained on either simple or complex traits, performance differed significantly across models trained on traits that differed in their level of correlation with one another (Simple: Kruskal-Wallis chi-squared = 52.459, p < 0.001, Complex: Kruskal-Wallis chi-squared = 52.459, p < 0.001). Models trained on uncorrelated traits performed best, followed by models trained on positively correlated traits, followed by models trained on negatively correlated traits.

**Conversation length.** Overall, the performance of models trained on simple traits was significantly better than the performance of models trained on complex traits (Wilcoxon rank sum test: W = 40000, p < .001), which can be seen in Figure 4. Within models trained on each trait complexity type, conversation length significantly impacted model performance (Simple: Kruskal-Wallis chi-squared = 68.601, p < 0.001, Complex: Kruskal-Wallis chi-squared = 36.849, p < 0.001). For models trained on both simple and complex traits, models trained on conversations with fewer turns tended to perform worse than models trained on conversations with greater numbers of turns. For simple traits, models trained on 50 turns performed significantly worse than models trained on all other conversation lengths. Models trained on 100 turns performed significantly worse than models trained on conversations with 300 or more turns, and models trained on conversations with 150 turns performed significantly worse than models trained on conversations with 400 or 500 turns. For complex traits, models trained on 50 turns performed significantly worse than models trained on 200 turns and 300 or more turns, models trained on 100 turns performed significantly worse than models trained on 300-450 turns, and models trained on 150 and 200 turns performed significantly worse than models trained on 400 turns (Figure 4).

**Group size.** Figure 5 visualizes the comparison of model performance as group sizes vary. Overall, the performance of models trained on simple traits was significantly better than the performance of models trained on complex traits (Wilcoxon rank sum test: W = 160000, p < .001). Within models trained on each trait complexity type, group size significantly impacted model performance (Simple: Kruskal-Wallis chi-squared = 15.144, p < 0.01, Complex: Kruskal-Wallis chi-squared = 9.2818, p < 0.05). For the models trained on simple traits, groups with 8 and 10 speakers both performed significantly better than groups with 4 or 6 speakers (Figure 5). For the models trained on complex traits, groups with 4 speakers performed significantly better than groups with 10 speakers (Figure 5). These results indicate opposite effects of group size depending on the complexity of the traits, with simple traits being learned better in larger groups and complex traits being learned better in small groups. However, note that these performance differences across group sizes are quite small.

## Discussion

Unsurprisingly, the performance of a given model in predicting patterns of turn-taking depends upon the structure of the processes that generated the conversational turn-taking data. The ability to learn memory functions when such functions underlie the structure of conversational turn-taking is critical to performance. While the presence of a memory component in the model did significantly detract from model performance when no such effect of memory was present in the data, the relatively small difference in performance is unlikely to have a substantial effect on model performance. Interestingly, the ability to learn individual-specific $\pi$ values was beneficial to model performance even when such individual differences were not present in the data. Thus, the use of our full model offers substantial benefits to model performance, regardless of the underlying processes that generated the data, with only minimal drawbacks.

The complexity of the functions underlying the relationships between an individual's traits and their speaking parameters impacted the difficulty of learning these functions, as expected. Thus, models will have greater predictive ability when an individual's traits are related to their speaking behaviors in a straightforward manner. Nevertheless, as demonstrated in Study 1, our model performs well even when learning complex functions.

Regardless of the level of complexity underlying the relationships between individual traits and model parameters, conversation length is an important variable to consider when selecting datasets that can be used with our model, particularly when conversations are less than 150 turns, and especially when they are less than 50 turns. Floor states, where two team members exchange speaking turns sequentially, may limit the number of opportunities that others outside of the floor have to speak, especially for very short conversations, thereby impeding the process of learning their speaking parameters. Thus, focusing on longer conversations may ensure that all team members have had the opportunity to express their unique speaking patterns.

As opposed to conversation length, how team members are divided across teams does not appear to substantially impact model performance. However, there was some evidence that smaller teams offer an advantage when trying to learn complex relationships between team member traits and their speaking behaviors and larger teams offer an advantage when trying to learn more simple relationships. One possible reason for why smaller teams may improve the learning of complex relationships is because each team member would have more opportunities to speak (and for the model to learn their behavior) than if they were part of a larger team. However, this explanation would presumably also apply to contexts where the relationships between individual traits and speaking behaviors are simpler. Thus, it remains unclear why we found opposing results for these two contexts. Nevertheless, differences in performance across different group sizes were small and unlikely to have a meaningful impact on model performance.

## Study 3

In Study 3, we applied our model to real-world data to demonstrate our model's ability to learn associations between team member characteristics (personality, gender) and their speaking behaviors in an authentic dataset. In addition, we conducted exploratory analyses focused on evaluating which characteristics or combinations of characteristics are most predictive of conversational turn-taking patterns and test predictions for how these characteristics relate to an individual's speaking behaviors.

### Research Questions and Predictions

The research questions we pursue in our third aim are as follows: Research Question 1) Which trait or combination of traits best predicts patterns of conversational turn-taking? Research Question 2) For our best-performing model, how do individual traits, or the joint effects of multiple individual traits, relate to an individual's speaking behaviors?

The traits we examined were extraversion, agreeableness, conscientiousness, openness, emotional stability, dominance, and gender. We selected these traits due to their associations with behaviors related to team communication. Extraversion is the personality trait that is most frequently linked to patterns of communication. For example, extraversion is positively correlated with verbal participation (McLean & Pasupathi, 2006) and talkativeness is often used as a measure of extraversion (Grant et al., 2011; McNiel & Fleeson, 2006; Wacker & Smillie, 2015; Watson & Clark, 1997). However, the other Big-5 personality traits have also been tied to communication behaviors. For example, those who are higher in agreeableness are thought to promote information sharing by others, rather than themselves. This has been proposed to be a reason why agreeableness is associated with team, but not individual, performance (Bradley et al., 2013; Graziano et al., 1996). Meta-analytic evidence also indicates that extraversion, conscientiousness, openness to experience, and emotional stability are all predictors of leader emergence (Ensari et al., 2011; Judge et al., 2002). Since one of the most consistent predictors of leader emergence is verbal participation (i.e., speaking frequently) (Chaffin et al., 2017; Mullen et al., 1989; Stein & Heller, 1979), all of these traits have the potential to predict speaking likelihood. Furthermore, dominance level has been found to positively correlate with total speaking time (Mast, 2002). Finally, results from meta-analyses indicate that there is a significant correlation between gender and verbal participation, with men speaking more often (Badura et al., 2018; Leaper & Ayres, 2007). *We predict a positive correlation between all traits and speaking likelihood, except for agreeableness, for which we predict a negative correlation. We also predict that males will have a higher speaking likelihood than females.*

Predictions become more difficult when individuals possess some traits that may promote communication (e.g., high extraversion) and some that may inhibit it (e.g., high agreeableness). We currently have limited understanding of how multiple traits interact within individuals to impact their behaviors. For example, while dominance positively correlates with total speaking time (Mast, 2002), in mixed-sex pairs, men talk more than women regardless of their relative dominance (Kimble & Musgrove, 1988). Thus, not only do we examine the ability of individual traits to predict speaking patterns, but also trait combinations.

**Design and Method**

*Dataset*

The dataset used in this study is described in detail in [citation to be added after review]. This dataset was collected as part of a set of studies focused on understanding the relationship between team member characteristics, speaking patterns, and team member effectiveness. Data collection took place in Summer 2020 and Spring 2021 from students enrolled in a University in the southern United States. Due to the Covid-19 pandemic, data were collected from student teams collaborating remotely via Zoom throughout the semester while working on engineering and data science course projects. The original dataset consists of 112 team members across 21 teams. However, we removed teams containing team members whose personality traits were unavailable. Thus, our dataset consists of 106 team members across 20 teams with a median (IQR) team size of 5 (4-6) team members. The personality traits of agreeableness, conscientiousness, emotional stability, extraversion, and openness were measured for all team members at the beginning of data collection using 10 items each from the International Personality Item Pool (IPIP) markers for the Big-Five factor structure (Goldberg, 1992). Dominance was measured using 11 items from the California Psychological Inventory (CPI) (Gough, 1996). Items were rated on a scale of 1 (very inaccurate) to 5 (very accurate). The median (IQR) number of speaking turns observed by each team was 2180 (1303 - 3439). Unlike the synthetic data, team members interacted over the course of multiple separate meetings, with a median (IQR) of 10 (6.75 - 13.5) meetings per team. Another difference from the synthetic data was that not all team members were present in all meetings. The median (IQR) number of team members missing from each meeting was 0 (0 - 1). Nevertheless, our model can easily handle these complexities.

*Procedure*

We divided the 20 teams into 12 training groups, 4 validation groups, and 4 test groups. We generated 20 different data trials with different combinations of teams falling into the training, validation, and testing groups for each trial. To do this, we randomized the order of the teams and assigned the first 12 to the training dataset, the next 4 to the validation dataset, and the final 4 to the testing dataset. We then slide forward these demarcations by 1, with the training dataset now including teams 2-13, the validation dataset including teams 14-17, and the testing dataset containing teams 18-20 and team 1. We continued sliding these demarcations forward by 1 team for each data trial. We trained our model on each of the individual-level traits described above for each data trial and calculated the ability of each trained model to predict the speaking sequences observed in the testing datasets (i.e., loss). Based on the performance of individual traits, we explored trait combinations that may better explain speaking patterns than each trait on its own. Following the selection of our best-performing model, we compared model performance to the performance of the same baseline models tested in Study 1 (but this time trained on real-world data). Finally, we visualized the relationships between the traits included in the best-performing *ML-SPEAK* model and model parameters to gain insight into the role individuals' traits played in their speaking patterns.

**Analysis**

We used a forward selection approach to evaluate which personality traits should be included in our final model. We first found the trait that performed best individually, then the best pair of traits that contained the initial best-performing trait, and finally the best triad of traits that contained the best-performing pair of traits. In order to promote interpretability, we limited our consideration of trait interactions to a maximum of three traits. We began with the same traits model as our baseline model, which did not consider individual variation in trait values. Then we evaluated uni-variate, bi-variate, and tri-variate models, testing how each successive increase in model complexity compared to the best-performing model that came before it (by subtracting each new model's loss value for each data trial from the previous best-performing model's loss value for that data trial). With each increase in model complexity, we considered all models whose median loss difference value was less than zero, indicating greater median performance compared to the baseline model used for comparison. If more than one model had better median performance compared to the baseline model, we selected the model with the lowest p-value. We calculated p-values using a paired, one-sided Wilcoxon Rank sum test (pairing loss values obtained from each data trial).

We further evaluated the performance of our best-performing model by comparing it to the same model baselines used in Study 1 (i.e., the *SPEAK*, *randomized traits*, *same traits*, *linear regression*, and *same traits, no memory* models). Following our procedure in Study 1, we took the loss values obtained by a given model for a given data trial and subtracted the loss value obtained for that same data trial by the *same traits* model. Thus, models with a loss difference less than zero indicate that it performed better than a model that did not consider individual variation in trait values. Although our model is capable of handling datasets consisting of conversations where not all team members are present at every meeting, our baseline models adapted from linear regression analysis and the original model by Stasser and Taylor (1991) are incapable of handing these inconsistencies. In order to enable fair comparison across models, we restricted this analysis to only conversations where all team members were present, resulting in a median (IQR) of 1333 (662 - 2526) speaking turns per team.

We visualized the relationships between the traits included in our best-performing *ML-SPEAK* model and model parameters $\pi$, $d$, as well as overall speaking likelihood $\ell_i\left(t_i^{\text{last}} + 2\right)$. Each visualization represents the average relationship learned by the *ML-SPEAK* model across all data trials. To plot the relationship between an individual trait and speaking behavior, we fixed all traits but one to their mean values in the dataset, and we plotted $\pi$, $d$, and $\ell_i\left(t_i^{\text{last}} + 2\right)$ with respect to the single trait. We also plotted the learned relationship for each data trial as lighter gray lines. As our best-performing model consisted of three traits, we visualized the learned relationship between two traits and speaking behavior via surface plots with respect to the two traits of interest, where we fixed the third trait to its minimum or maximum value in the dataset.

**Results**

We examined uni-variate model performance by evaluating the difference in loss between a model trained on each individual trait and the *same traits* baseline, which did not consider individual differences (Figure 6). Extraversion, emotional stability, and openness all had median loss values less than 0, indicating greater median performance compared to the *same traits* baseline. When comparing the performance of these uni-variate models to the performance of the same traits model, the model including extraversion had the lowest p-value and was the only model that predicted speaking turn sequences significantly better than the *same traits* baseline (W = 58, p < 0.05), with emotional stability (W = 84, p = 0.23) and openness not reaching significance (W = 121, p = 0.73).

We examined bi-variate model performance by comparing models trained on extraversion combined with each of the other individual traits to the uni-variate model trained on extraversion alone. Of the models examining pairs of traits, models trained on extraversion and agreeableness, openness, conscientiousness, and emotional stability all had median loss difference values less than 0, indicating greater median performance compared to the *extraversion* model (Figure 6). The model trained on extraversion and agreeableness had the lowest p-value and was the only model that performed significantly better than the *extraversion* model (W = 58, p < 0.05), with models combining extraversion with emotional stability (W = 74, p = 0.13), openness (W = 91, p = 0.31), and conscientiousness (W = 94, p = 0.35) not reaching significance.

We examined tri-variate model performance by comparing models trained on extraversion and agreeableness combined with each of the other remaining individual traits to the bi-variate model trained on extraversion and agreeableness. Of the models examining three traits, the model trained on extraversion, agreeableness, and emotional stability was the only model with a median loss difference value less than 0, indicating greater median performance compared to the *extraversion and agreeableness* model (Figure 6). However, the difference in performance between the *extraversion, agreeableness, and emotional stability* model and the *extraversion and agreeableness* model was not significant (W = 91, p = 0.31).

For our subsequent analyses, we focused on the model that had the best median level of performance, which was the *extraversion, agreeableness, and emotional stability* model. When comparing the performance of this model to the model baselines used in Study 1, model type had a significant effect (Kruskal-Wallis chi-squared = 75.87, p < 0.001) on the ability to predict speaking turn sequences (i.e., loss value). Our *ML-SPEAK* model trained on extraversion, agreeableness, and emotional stability performed significantly better than the other baseline models in predicting speaking sequences (Figure 7) and was the only model that performed better than the *same traits* baseline. All models performed significantly differently from one another except for the *linear regression* and *same traits, no memory models*. As in the model comparisons using the synthetic data (Study 1), all models with the memory component performed better than the models without the memory component. However, in contrast to model comparisons using the synthetic data, the *SPEAK* model performed worse than both the same traits model and the randomized traits models.

**Figure 6**

*Results of Forward Selection*

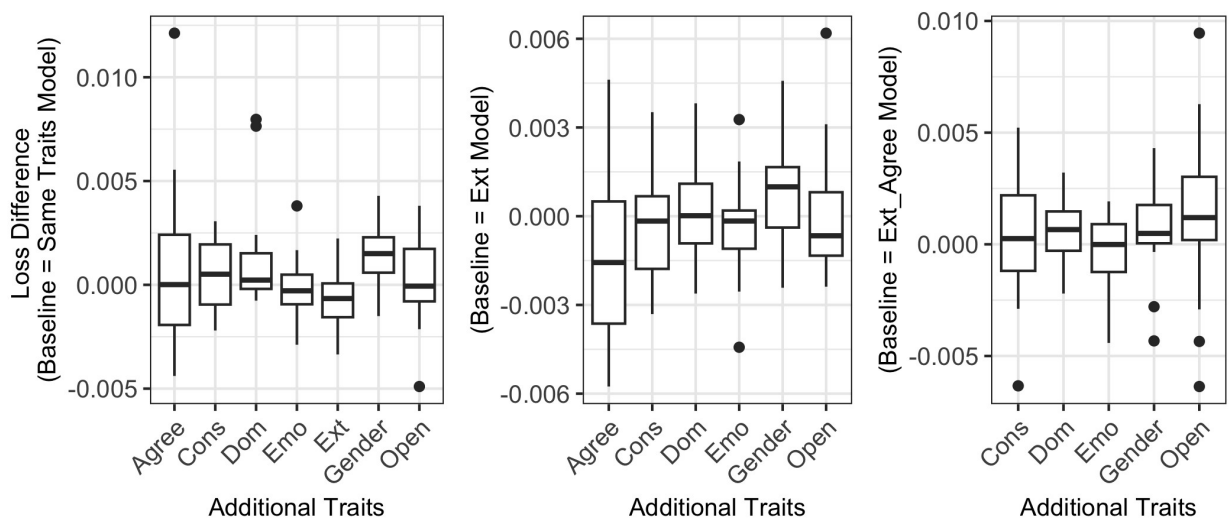

*Note.* Comparison of model performance when trained on individual traits and trait

combinations. $y = 0$ indicates the loss obtained for a given data trial by the A) same traits model,

B) the model trained on extraversion only, and C) the model trained on both extraversion and

agreeableness. These are the baseline models to which the results of each panel are compared.

The traits listed on the *x*-axis represent the traits taken into account in addition of those

considered by the baseline model listed on the *y*-axis. "Agree" = agreeableness, "Cons" =

conscientiousness, "Dom" = dominance, "Emo" = emotional stability, "Ext" = extraversion,

"Open" = openness.

**Figure 7**

*Model Performance Comparisons for Real Data*

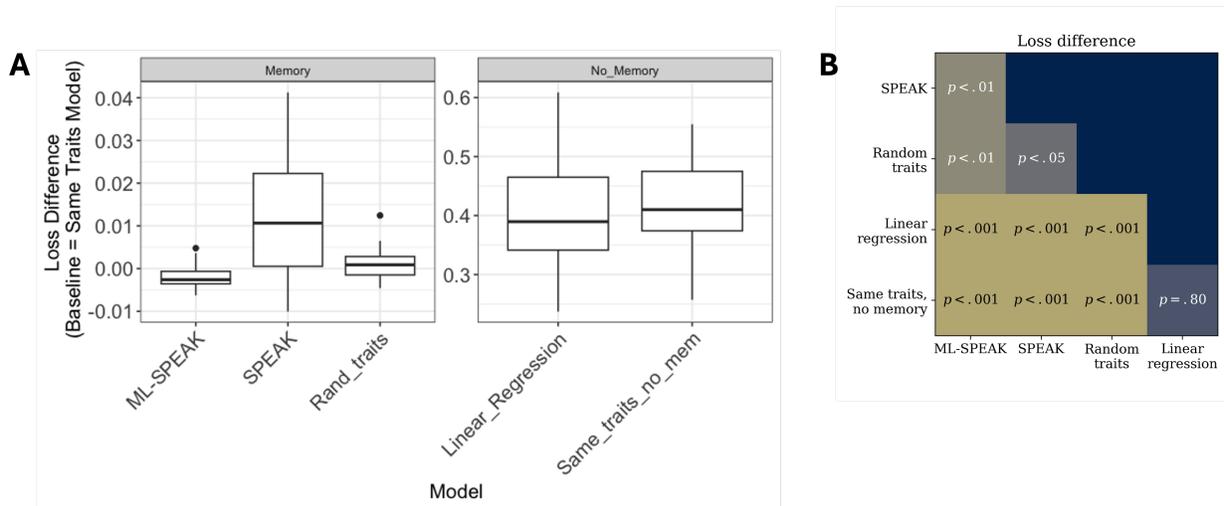

*Note.* "Loss Difference" results represent the difference in loss between each model and the *same traits* model with lower values indicating better performance. A) "ML-SPEAK" corresponds to our proposed model trained on the traits extraversion, agreeableness, and emotional stability, while the baseline models used for comparison are denoted in the figures by "Same_traits", "Same_traits_no_mem", "Rand_traits", "Linear_Regression", and "SPEAK", respectively. B) Results of Wilcoxon multiple comparisons test comparing performance between each pair of models.

**Figure 8**

*Visualization of Learned Relationships for Individual Traits*

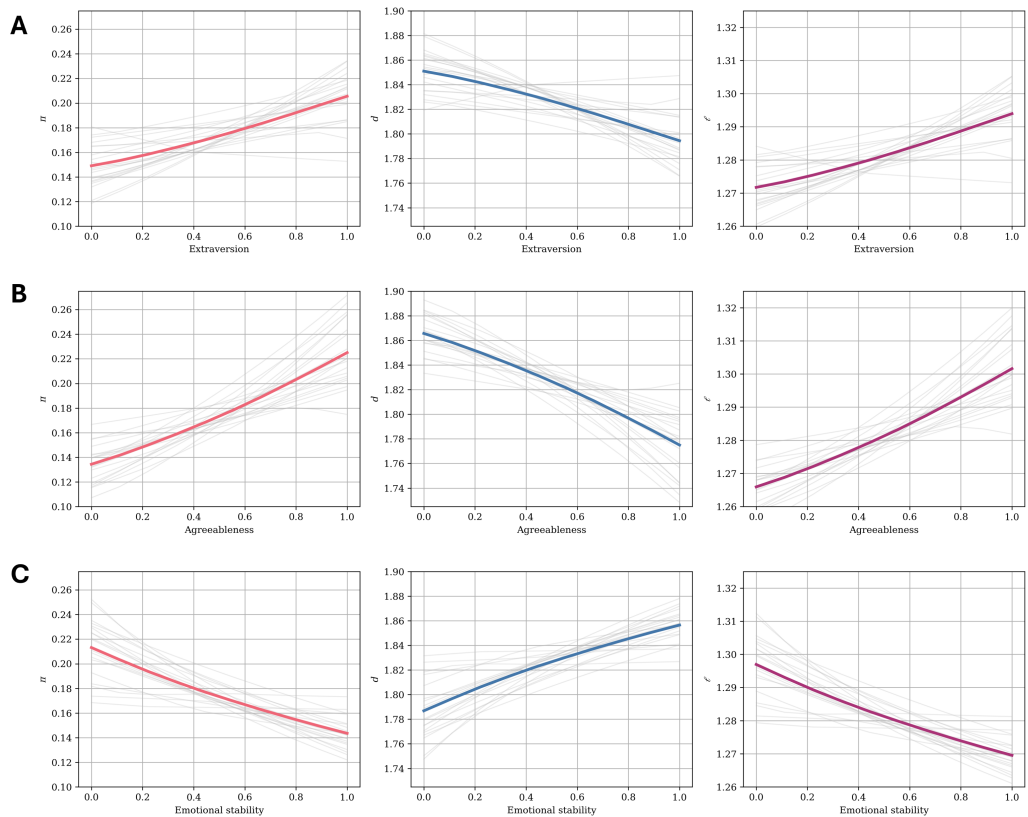

*Note.* Plots representing the mean relationships learned between individual traits and $\pi$, $d$, and speaking likelihood. The colored/darker line in each plot represents the mean learned relationship with the lighter colored lines representing the relationship learned in each data trial. A plots the learned relationship for extraversion while holding agreeableness and emotional stability at their mean values. B plots the learned relationship for agreeableness while holding extraversion and emotional stability at their mean values. C plots the learned relationship for emotional stability while holding extraversion and agreeableness at their mean values. The *x*-axes represent normalized trait values, where 0 denotes the smallest value of the trait observed in the dataset and 1 the highest.

**Figure 9**

*Visualization of Learned Relationships for Trait Combinations*

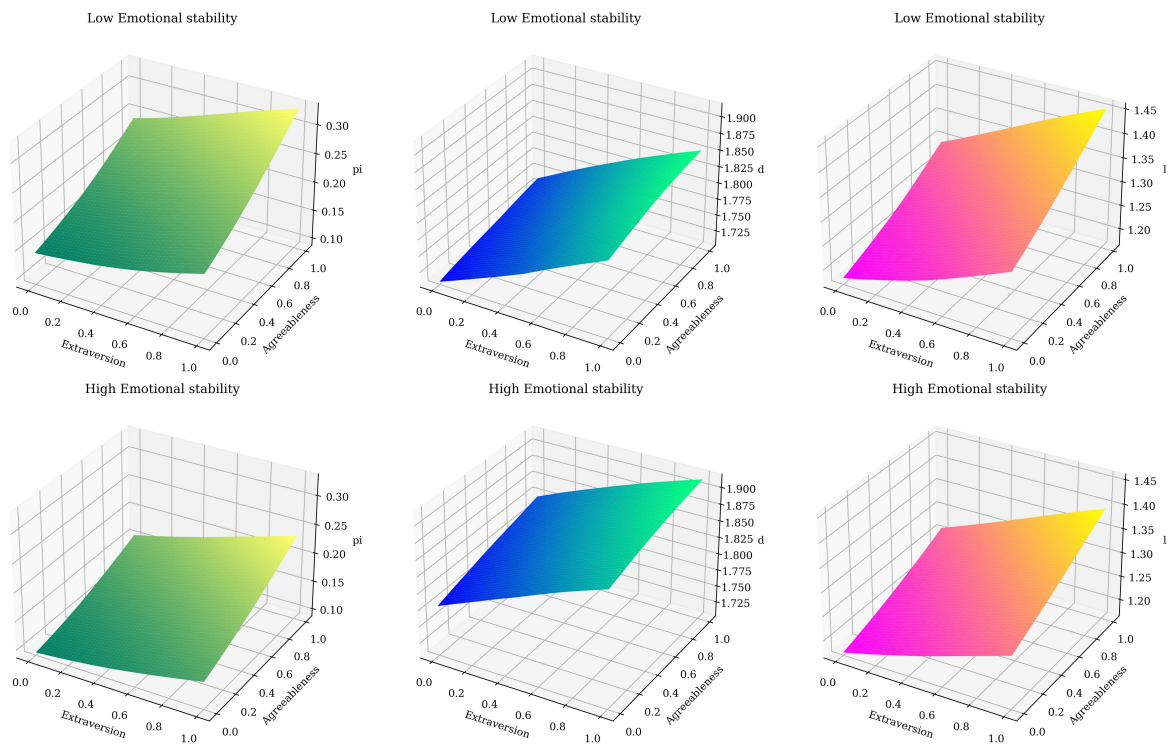

*Note.* Surface plots representing the mean relationships learned between extraversion and agreeableness (at two different levels of emotional stability) and $\pi$, $d$, and speaking likelihood. The axes labeled "extra" and "agree" are normalized trait values, where 0 denotes the smallest value of the trait observed in the dataset and 1 the highest.

Finally, we visualized the learned relationships between the speaking behaviors built into our model and the personality traits extraversion, agreeableness, and emotional stability. Figure 8 displays the mean learned relationships between each of the three traits included in the best-performing model (while holding each of the other two traits at their mean value) and $\pi$, $d$, and overall speaking likelihood $\ell_i\left(t_i^{\text{last}} + 2\right)$. Extraversion and agreeableness had positive relationships with baseline likelihood of speaking ($\pi$) and negative relationships with an individual's tendency to speak after having recently spoken ($d$) (reflecting level of interactivity) while emotional stability displayed the opposite relationships. Both extraversion and agreeableness had a positive relationship with overall likelihood of speaking ($\ell_i\left(t_i^{\text{last}}+2\right)$), and emotional stability had a negative relationship with overall likelihood of speaking. These results for overall likelihood of speaking support our prediction for extraversion but display the opposite relationships to those predicted for agreeableness and emotional stability.

Figure 9 shows the surface plots displaying the relationships between the same speaking behaviors examined above and the interaction between an individual's level of extraversion and agreeableness, at either low or high levels of emotional stability. Overall, we find that baseline speaking likelihood ($\pi$) primarily increases with an individual's level of agreeableness, with a small positive effect of extraversion, while an individual's tendency to speak after having recently spoken ($d$) primarily increases with their level of extraversion. Baseline likelihoods of speaking were overall higher when individuals were lower in emotional stability, but their level of interactivity (controlled by $d$) was higher when they were higher in emotional stability. Taken together, both extraversion and agreeableness were positively associated with an individual's overall speaking likelihood, with a small negative effect of emotional stability.

**Discussion**

The results of Study 3 show that personality traits demonstrate predictive ability for speaking behavior. The model trained on extraversion was the only uni-variate model that predicted speaking patterns significantly better than a model that did not consider any individual differences in speaking behavior. This finding is consistent with abundant research on the relationship between extraversion and an individual's communication and social behaviors (Grant et al., 2011; McLean & Pasupathi, 2006; McNiel & Fleeson, 2006; Wacker & Smillie, 2015; Watson & Clark, 1997), validating the ability of our model to learn this well-documented relationship. Moreover, we found that the combination of the most informative traits, extraversion, agreeableness, and, to a lesser degree, emotional stability, provided even greater predictive performance. These results show that combinations of personality traits can be more informative than individual traits alone, highlighting the intricacy of their effect on individual behavior. As the Big-5 personality traits are not necessarily orthogonal (Ones, 2005), that is, the variation of one may be correlated with another, the influence of multiple traits may not be merely additive (Witt et al., 2002). Indeed, extraversion and emotional stability have been deemed informative for individual performance when considered together, whereas either trait separately is less associated with performance (Judge & Erez, 2007).

We did not find substantial non-linearities in the relationships between extraverison, agreeableness, and emotional stability and model parameters. Nevertheless, our model is significantly better at predicting speaking patterns compared to linear regression models which can estimate the linear relationships between individual traits and overall speaking probabilities. This is likely due to our model's ability to take into account the time-dependent nature of turn-taking behavior and to break down speaking likelihood into both stable ($\pi$) and sequence-dependent ($d$) components. Furthermore, our model retains the ability to learn more complex relationships between traits and speaking behaviors, when present, as demonstrated in Study 1.

In contrast to our predictions, agreeableness displayed a positive relationship with overall speaking likelihood while emotional stability displayed a negative relationship. Agreeableness has been associated with promoting information sharing by others (Bradley et al., 2013; Graziano et al., 1996), and our model indicated that this trait was associated with a higher baseline likelihood of speaking, but a lower level of interactivity. One potential explanation for these findings is that more agreeable individuals prompt others to speak, but do not get wrapped up in long turn-taking interactions themselves. Interestingly, this is the same pattern displayed by individuals low in emotional stability. As most research on communication behaviors does not break down individuals' tendencies to speak into both stable and sequence-dependent speaking behaviors, more work is needed to develop theories regarding how individual differences may relate to these different ways of engaging during conversation.

## General Discussion

This work demonstrates the potential for personality traits to predict communicative behavior, not only when analyzing how an individual will tend to speak but also how a conversation among a group of individuals will tend to evolve. Our findings in Study 1 show that our proposed method, which combines existing conversational models with powerful predictive tools, can predict patterns of conversational turn-taking with superior performance to existing approaches, including the original model by Stasser and Taylor (1991). Our model also demonstrates the unique ability to learn the association between these patterns of conversational turn-taking and individual differences between team members.

Our results in Study 2 reveal how our model learns under different conditions, demonstrating the settings under which we can expect greatest performance. As would be expected, the amount of conversational data among groups must be sufficient to relate speaking behavior to personality, particularly as the relationship between individual differences and speaking behaviors becomes more complex. However, we do not observe as much of an effect due to group size, supporting the application of our model in Study 3 to real-world data that consists of multiple groups of different sizes. We should emphasize that, opposed to other methods such as linear regression and the original *SPEAK* model by Stasser and Taylor (1991), our model is not hindered from learning and predicting speaking behaviors in teams that vary in size and composition both across and within teams (i.e., when not all team members are present in a given conversation). This is a clear advantage for working with real-world datasets. Moreover, we found that our standard *ML-SPEAK* model, which includes the ability to learn individual differences in both baseline speaking likelihood and turn-dependent speaking likelihood, performs well across datasets regardless of the role individual differences or turn-dependent speaking behaviors play in generating those data.

Our findings in Study 3 address our two posed research questions. First, we found that extraversion, agreeableness, and emotional stability were the three traits among those considered that demonstrated the greatest influence on speaking behavior in our dataset. While the results of our study were exploratory, they are in line with the abundant literature on the positive association between extraversion and communication behaviors (Grant et al., 2011; McLean & Pasupathi, 2006; McNiel & Fleeson, 2006; Wacker & Smillie, 2015; Watson & Clark, 1997). Our results are also consistent with studies that have determined that taking into account the interaction between multiple traits can be more informative than considering these traits individually (Judge & Erez, 2007; Lynn & Steel, 2006; Witt et al., 2002). For example, our finding that extraversion and agreeableness together are significantly more informative than either alone aligns with this view and parallels existing findings that the combination of these two traits is more indicative of job performance than either alone (Judge & Erez, 2007).

Beyond the verification of existing theory, our model has advantages over existing methods for describing conversational behavior because of its ability to perform predictive analyses. Indeed, the classical model in Stasser and Taylor (1991) cannot relate personality traits to speaking behavior. Learning how personality traits yield differences in communicative behavior is complex, not only due to the complex ways in which these traits may combine within an individual to influence behavior but also due to the complex interactions between the traits of multiple interacting individuals. Not only do we address this complexity by leveraging machine learning methods, but we also mitigate some of the concerns about the lack of interpretability of deep learning (Stachl et al., 2020). By using a domain-inspired model, we can exploit powerful machine learning tools that can learn complex patterns that may be imperceptible under traditional statistical analyses while allowing for some interpretation between personality traits and speaking tendencies. For example, in Figure 9, we can observe how the model predicts an individual's speaking likelihood as personality traits vary without requiring any observed conversations.

**Limitations and Future Directions**

While our study showed promise for extracting meaningful relationships between the personality traits of team members and their tendencies to speak, our analysis of real-world data was restricted to a single dataset with a relatively small number of teams. Moreover, these conversations were observed over Zoom. While video conferencing technology can deliver greater media richness than other types of computer-mediated communication (Purdy et al., 2000), personality may be less influential on behavior in virtual conversations (Bradley et al., 2013). Furthermore, research suggests that team members using video conferencing technology tend to exhibit more competitive behavior and operate less efficiently compared to those interacting in person (Purdy et al., 2000). Thus, in order for our model to promote the development of new theory on the relationships between team member traits, team composition, and conversation dynamics, it must be applied to larger datasets collected through a variety of mediums.

Our study also only considered a relatively small subset of traits that may impact speaking behaviors. Additional combinations of traits not considered here may be found to be more informative of speaking behavior. Thus future work should consider a broader range of individual characteristics, including those such as abilities, motivation, or language proficiency (Bottger, 1984; Li et al., 2019b; Littlepage et al., 1995). For example, during decision-making tasks, confidence in one's ideas has been found to positively correlate with speaking time (Littlepage et al., 1995). Furthermore, the characteristics that are most predictive of communication patterns have the potential to change based upon the task at hand, type of team, or context of interaction. Thus, extensive work is needed to understand contextual variation in the link between individual differences and communication patterns.

In terms of the inner workings of our model, the restriction of speaking likelihood to the form in (1) may limit the model's ability to learn more complex speaking patterns. For example, some individuals may experience a decaying effect of speaking likelihood as more turns occur (Stasser and Taylor, 1991) as assumed by our model, while others may have a higher chance of speaking the longer that they do not speak. Thus, future work could replace the existing memory function (which imposes a pre-defined decay rate in speaking likelihood) with another learning-based model.

Furthermore, future modifications to the model can allow the learning of more nuanced relationships between individual differences and speaking behaviors. For example, our current model assumes that individual traits have a constant relationship with speaking behaviors and team dynamics. However, team processes often evolve over time as the team shifts their focus to different tasks and these shifts can impact the role individual differences play in team dynamics (LePine, 2003; Marks et al., 2001). Thus, future extensions can add time-varying parameters to enable a more dynamic investigation into team communication patterns over time. We expect that notions from transfer learning and fast retraining can be useful in this research direction (Lu et al., 2015; Tan et al., 2018).

## Contributions and Broader Implications

We provide a computational model that learns how speaking behavior is affected by personality traits and allows us to predict how teams will communicate without any prior observations of conversations between team members. Using this predictive approach, we can provide powerful insights into the relationships between individual traits, team composition, and team communication processes. For example, our trained model can yield interpretable, quantitative results for a greater understanding of why certain communication patterns arise. Thus, our method has the potential to inform theories of team processes that seek a mechanistic understanding of emergent team communication dynamics (Kozlowski & Chao, 2018; O'Bryan et al., 2020). Moreover, as communication patterns have been linked to many team outcomes (Engel et al., 2014; Sherf et al., 2018; Woolley et al., 2015), this functionality has the potential to support understanding of why some team compositions perform better than others.

By better understanding how team processes emerge from the interaction between team member traits and behaviors, we have the potential to design teams in a manner that will produce favorable team processes or design interventions to improve team processes once they develop. Thus, our model can become a powerful tool for designing human teams. Applications developed from our work can enable working backward from desired communication patterns to determine the team compositions that would most likely lead to productive team-level communication processes. Moreover, by predicting the conversational interactions among team members, one can anticipate undesirable outcomes in pre-established teams and apply interventions ahead of time. For example, if our model predicted that a given set of team members would form one or more subgroups, actions could be tailored toward mitigating this tendency. Similarly, for established teams, our tool could be used to predict the changes in communication patterns that may occur due to group composition changes, thus providing guidelines for restaffing or retraining team members. Finally, should task demands change (e.g., the need to shift from a strongly centralized team to an egalitarian team), our model could help to quickly determine the personnel changes or training interventions that should be utilized to best achieve that goal. Thus, by improving our understanding of how team processes develop, our proposed model has the potential to enable more precise control over team development, training, and modification.

## Conclusion

In this work, we presented a computational model that relates an individual's personality traits to their conversational turn-taking behavior. Furthermore, our model can combine the traits of a group of individuals to predict the patterns of communication expected from a conversation among those individuals. Our model was designed to extract speaking tendencies from observed conversations, revealing individuals' idiosyncratic conversational behavior with respect to their personality traits, despite the complexities of observing their behavior in a group setting where all team members' speaking behaviors have the potential to influence one another. The ability of domain-inspired deep learning tools to provide predictive and intuitive results for personality-focused research allows the potential for gaining novel insight into complex behavioral patterns. As traditional statistical approaches may be insufficient for such insights, we show that they may be attainable without the need for "black box" models that lack interpretability. Our work contributes to this necessary step forward, and our findings demonstrate that there is much to be discovered with such an approach.

## Acknowledgements

This work was funded by the Army Research Institute for the Behavioral and Social Sciences [Grant number to be added after review]. The views, opinions, and/or findings contained in this paper are those of the authors and shall not be construed as an official Department of the Army position, policy, or decision, unless so designated by other documents.